\documentclass{svproc}
\usepackage{url}
\usepackage{graphicx}   % Required for \includegraphics
\usepackage{hyperref}   % Required for \url
\usepackage{lipsum}     % For placeholder text (remove if not needed)
\usepackage{amsmath}
\usepackage[numbers]{natbib} % Use natbib for numbered references

\begin{document}
\mainmatter              % start of a contribution
\title{Evaluating Compositional Approaches for Focus and Sentiment Analysis}
\titlerunning{Evaluating Compositional Approaches}  % abbreviated title (for running head)
\author{Olga Kellert\inst{2} \and Muhammad Imran\inst{1}\thanks{Corresponding author: Muhammad Imran (m.imran@udc.es)} \and
Nicholas Hill Matlis\inst{3} \and Mahmud Uz Zaman\inst{4} \and Carlos Gómez-Rodríguez\inst{1}}
\authorrunning{O. Kellert et al.} % abbreviated author list (for running head)
\tocauthor{First Author, Second Author, Third Author and Fourth Author}
\institute{Universidade da Coruña, Grupo LyS, CITIC, Depto. de Ciencias de la Computación y Tecnologías de la Información, Campus de Elviña s/n, 15071 A Coruña, Spain,\\
%\email{{o.kellert@udc.es} {m.imran@udc.es} {carlos.gomez@udc.es}}
%\email{\{{o.kellert | m.imran | carlos.gomez}\}@udc.es}
\and
School of International Letters \& Cultures,
Arizona State University, Tempe, Arizona, United States
\and
Beus CXFEL Laboratory, Biodesign Institue \& Physics Department, College of Liberal Art \& Sciences, Arizona State University, Tempe, Arizona, United States
% \and
% Center for Free-Electron Laser Science CFEL, Deutsches Elektronen-Synchrotron DESY, Germany
\and
University of Augsburg, The Applied Computational Linguistics (ACoLi) Lab, Universitätsstr. 10, 86159 Augsburg, Germany}

\maketitle              % typeset the title of the contribution
\begin{abstract}
  This paper summarizes the results of evaluating a compositional approach for Focus Analysis (FA) in Linguistics and Sentiment Analysis (SA) in Natural Language Processing (NLP). While quantitative evaluations of compositional and non-compositional approaches in SA exist in NLP, similar quantitative evaluations are very rare in FA in Linguistics that deal with linguistic expressions representing focus or emphasis such as "it was {\itshape John} who left". We fill this gap in research by arguing that compositional rules in SA also apply to FA because FA and SA are closely related meaning that SA is part of FA.       
  Our compositional approach in SA exploits basic syntactic rules such as rules of modification, coordination, and negation represented in the formalism of Universal Dependencies (UDs) in English and applied to words representing sentiments from sentiment dictionaries. Some of the advantages of our compositional analysis method for SA in contrast to non-compositional analysis methods are interpretability and explainability. We test the accuracy of our compositional approach and compare it with a non-compositional approach VADER that uses simple heuristic rules to deal with negation, coordination and modification. In contrast to previous related work that evaluates compositionality in SA on long reviews, this study uses more appropriate datasets to evaluate compositionality. In addition, we generalize the results of compositional approaches in SA to compositional approaches in FA.
% We would like to encourage you to list your keywords within
% the abstract section using the \keywords{...} command.
\end{abstract}
\section{Introduction}
Focus has an important role in Natural Languages. It helps to disambiguate sentences and clarify the speaker's intent. By emphasizing a particular element in a sentence, speakers can guide listeners to the intended interpretation. For example, in the sentence "{\itshape John} left" or " It was {\itshape John} who left" the focus on "John" clarifies that it is John, not someone else, who left. This helps avoid misunderstandings and ensures the message is conveyed accurately. There are various theories in linguistics that address how focus interpretation is expressed or derived in natural languages, and these theories can be categorized based on their stance on compositionality. Below is an overview of some key theories, distinguished by whether they consider focus meaning to be derived compositionally or not:

\begin{itemize}

\item \verb|Compositional theories| : Alternative Semantics \cite{rooth1985,beck2006,kellert2015}, Structured Meanings \cite{vonstechow1991,krifka1991}
\item \verb|Non-Compositional theories| : Contextual Theories \cite{roberts1996}, Cognitive Theories \cite{chafe1976,lambrecht1994}

\end{itemize}

In a compositional analysis of focus, the meaning of a sentence with a focused element is built up systematically from the meanings of its parts \cite{rooth1985,beck2006,kellert2015}. This approach relies on formal rules that ensure each component of a sentence contributes to the overall meaning in a predictable manner \cite{rooth1985,beck2006,jacobs1984,chierchia2004,kellert2015} among others. Rooth's focus theory is a prime example of a compositional analysis. It posits that every sentence has two parallel interpretations: 
\begin{itemize}

\item \verb|Ordinary Interpretation| : The standard meaning of the sentence, e.g. John left.
\item \verb|Focus Interpretation| : A set of alternatives that highlight the focus: {John left, Mary left, Peter left}

\end{itemize}

Each of these interpretations are derived using compositional rules, e.g. combining the meaning of "John" with the meaning of "left" using syntactic rules to a more complex meaning represented as a sentence "John left". This ensures that the meanings are built up from the parts "John" and "left" consistently, handling the ordinary and focus interpretations, respectively.

In a non-compositional analysis, the meaning of a sentence like "John left" is not strictly derived from its parts "John" and "left". Instead, the interpretation of "{\itshape John} left" or " It was {\itshape John} who left" depends on contextual or pragmatic factors that are not systematically predictable from the components of the sentence alone. The focus is understood primarily through its pragmatic impact \cite{roberts1996,chafe1976,lambrecht1994}.

So far, this theoretical discussion between compositional and non-compositional approaches for FA has not been tested quantitatively and/or automatically on a large dataset in linguistics. Instead, the theoretical discussion is usually evaluated qualitatively based on a few sentences illustrating problems for the compositional or the non-compositional analysis method see \cite{kellert2015} for an overview. 

Our first goal in this paper is to fill this gap in linguistic research and to provide a way to test compositional and non-compositional approaches for FA empirically and automatically based on a larger dataset than a handful of picked sentences. To do this, we will use available datasets and experimental approaches from a related field to FA, namely Sentiment Analysis (SA) in Natural Language Processing (NLP). SA in NLP deals with the automatic prediction of the polarity orientation of a sentence or paragraph such as positive or negative. For instance, the sentence "Chocolate is tasty" is associated with a positive orientation or polarity due to the word "tasty". We argue that the projection of the polarity orientation of words like "tasty" to the sentence "Chocolate is tasty" is related to the compositional rules of how focus or emphasis projects to the sentence level as in "{\itshape John} left" or " It was {\itshape John} who left". This means that both projections (polarity and focus projection) are affected by the same syntactic rules such as coordination, modification, and negation  see \cite{rooth1985,krifka1991,beck2006,kellert2015} for syntactic rules of focus projection and see \cite{Vilares2017, kellert2023experimenting} for polarity projection in SA). To adapt approaches from SA to FA, it is necessary to modify the evaluation methods usually used to evaluate compositional and non-compositional approaches in SA \cite{Vilares2017, kellert2023experimenting}, because the evaluation is usually based on the accuracy prediction of reviews expressing sentiments, which can be very long. Predicting long reviews correctly requires more than compositional rules of how polarity projects to the sentence level considering syntactic rules of negation, modification and coordination. Instead, predicting long reviews also requires knowledge about how discourse relations between sentences work or to what extent stylistic rules matter for accuracy of polarity prediction. The latter rules are syntax or grammar-independent. For instance, some sentences in a long review play a more important role in polarity prediction such as first or last sentences. These grammar-external factors represent a "noise" factor in the evaluation of compositionality based on syntactic or grammatical rules expressed within a sentence. It is thus important to find a dataset with relatively short reviews (ideally one sentence long) for the evaluation of compositionality.  

Our contribution to this article is three-fold: 1) Find or create a dataset that targets compositionality and minimizes the influence of syntax-external factors, 2) Test compositional with non-compositional approaches on this dataset and 3) Generalize the results from SA to FA as both analyses are strongly related.

The article is structured as follows. Section 2 provides more theoretical details about compositional approaches of focus and sentiment analysis. Section 3 describes the modification of the previous compositional approach and a method for the data selection and Section 4 presents the results and discussion. Section 5 provides conclusion and section 6 provides limitations and future work.

\section{Compositional Approaches to Focus and Sentiment Analysis}

Rooth's \cite{rooth1985} focus theory is a foundational framework in semantics that addresses how elements within a sentence can receive special emphasis, or focus, and how this affects their interpretation. According to Rooth, focus elements have two distinct semantic interpretations: a) Ordinary Interpretation (Ordinary Semantic Value): This is the conventional meaning of the sentence without any emphasis. For example, in the sentence "{\itshape John} left," the ordinary interpretation is simply "John left." and b) Focus Interpretation (Focus Semantic Value): This captures the alternatives that could potentially replace the focused element within the sentence, reflecting the range of possible contrasts. In the same example, the focus interpretation might be the set {John left, Mary left, Peter left}, indicating that any of these individuals could have been the subject of the sentence. As one alternative is true, namely "John left", this leads to the inference that other alternatives are false. This explains why the focus marking on "John" often leads to the contrasting interpretation of the sentence ('It's John who left, not Mary or Peter'). Rooth's theory proposes that these two interpretations are computed in parallel. This dual interpretation framework allows for a nuanced understanding of how focus operates within a sentence, affecting both its meaning and its pragmatic use in discourse. In Rooth's \cite{rooth1985} focus theory, the interaction between focus and negation is particularly relevant as focus influences the interpretation of negated statements. When a focused element appears within a negated sentence, the ordinary and focus interpretations interact to produce the meaning of the sentence and its alternatives. For example, in the sentence "{\itshape John} did not leave," the ordinary interpretation is straightforwardly "John did not leave." However, the focus interpretation considers a set of alternatives such as {John did not leave, Mary did not leave, Peter did not leave}. This interaction can lead to nuanced readings, such as emphasizing that it was John, and not someone else from the alternative set, who did not leave. Coordination plays an important role in FA too \cite{rooth1985}. The coordination within the sentence with a focus element leads usually to a set of coordinated alternatives. The example "It was John and Mary who left" triggers the set of coordinated alternatives like {John and Mary left, John and Peter left, Peter and Susan left, ...} excluding non-coordinated alternatives like {John left (by himself), Mary left (by herself), ...}. Rooth's framework ensures that both the negation, coordination, and the focus are properly accounted for in deriving the meaning, preventing unintended interpretations, and maintaining coherence with the contextual alternatives. As discussed in appendix~\ref{subsec:formal_details_FA}, the formal details of FA are important for understanding the method.

Compositional approaches have been also implemented in SA such as \cite {Vilares2017, kellert2023experimenting}, which apply the principles of compositional semantics where the meaning of a sentence is derived from the meanings of its parts and the syntactic rules used to combine them. These approaches break down a review into sentences and each sentence into its syntactic components, represented as nodes in a tree structure. Each node captures a word, its context, and its syntactic dependencies, forming a hierarchical representation of the sentence's structure. The sentiment score is evaluated at each node based on the word and its context expressed in the corresponding node. This mirrors how human language processing works according to compositional analysis of Natural Languages (§Introduction), where the meaning of a phrase is understood by combining the meanings of individual words according to syntactic rules. The appendix~\ref{subsec:formal_details_SA} provides the formal details of SA.

The recursive traversal of each node in the hierarchical representation of a sentence allows compositional approaches to account for the context and syntactic dependencies that influence the polarity orientation of the sentence. Words in natural language often depend on their context to convey the correct sentiment. By analyzing the tree structure, compositional methods can effectively manage such dependencies. This ensures that each sentiment word's score is appropriately weighted and contextualized within the sentence.

Some of the advantages of a recursive and compositional approach of SA is that SA is comprehensive and explainable. By breaking down sentences into smaller, manageable parts and analyzing each part in its context, compositional approaches can aggregate individual word sentiments into a coherent overall score. This methods effectively capture the nuances and complexities of natural language, producing a more accurate and reliable sentiment analysis  \cite {gomez2024dancing}. Some compositional analyses in SA exploit words expressing sentiments such as "tasty" from sentiment dictionaries and syntactic rules together with sentiment-shifting elements like negation and modification as in "This cholocate is not very tasty" \cite {Vilares2017, kellert2023experimenting}. These approaches have used the formalism of Universal Dependencies (UD) which is a universal framework for the annotation of grammar across different human languages \cite {11234/1-5150} in order to capture syntactic rules of polarity projection \cite {Vilares2017, kellert2023experimenting}. The authors in \cite {kellert2023experimenting} tested their analysis on the dataset provided by the Shared Task Rest-Mex 2023 organizers \cite{alvarez2023overview} and compared the results of their compositional analysis with a comparable dictionary-based non-compositional analysis of SA that use heuristic rules to address modification and negation such as VADER \cite {Hutto_Gilbert_2014}. In addition, they compared their results with non-compositional and non-dictionary-based approaches based on Deep Learning methods. Their results have shown that their compositional approach is superior to VADER in the accuracy of polarity prediction of long reviews that can contain up to 20 sentences \cite {kellert2023experimenting}. While previous compositional approaches like the one from \cite {kellert2023experimenting} implemented negation and modification, they mostly ignored the relation between negation, coordination, and modification. In the next section, we show how we modify the code from \cite {kellert2023experimenting} to include more complex sentences such as coordination and we discuss the appropriate datasets we used to evaluate compositionality by reducing sentence external factors that might influence the accuracy prediction. 

\section{Data and Methods}
\subsection{Code Modification}
We suggest a modification of the code in \cite {kellert2023experimenting} that does not account for complex sentences combined by coordination. 
 
It specifically deals with the interaction between negation and coordination. 

In the modified version of \cite {kellert2023experimenting}, coordination has the scope over the negation predicting the correct interpretation of conjoined sentences. For instance, in the sentence "No es muy costoso pero tiene una vista bonita" ("It is not very expensive but it has a beautiful view"), the negation word "no" inverts the negative sentiment of "costoso,", but not of the sentiment word "bonita", contributing to a more accurate overall sentiment score. The link to the modified version of the code is available under § Online Resources.

\subsection{Dictionaries}

We use the sentiment dictionary SO-CAL for English \cite {BroTofTab2009a}. The content of this dictionary and its parameters are not modified or tuned. For comparison with the non-compositional method, we use the sentiment dictionary that is already built into the non-compositional approach \cite {Hutto_Gilbert_2014}. 

\subsection{Data}
We use a dataset of 1744 hotel reviews in English from OpeNER \cite {Agerri2013}. It was extracted from different booking sites from November 2012 to November 2013. Each review is annotated with individual polarity expressions and their polarity (positive or negative) as demonstrated by a simple example such as "My best honeymoon." (Polarity: Positive) from \cite{barnes-etal-2022-semeval}. The dataset has additional information such as polarity holders or agents, etc. \cite{barnes-etal-2022-semeval}, which we ignore. The mean count of sentences per review in the OpeNER dataset is 1.06. The mean count of tokens per sentence is 16.38, which means that the sentence approximately contains 16 words on average. This makes this dataset a good choice for our goal as we want to test how the polarity projects on the sentence level and reviews with several sentences pose an additional complication to this goal. Furthermore, we performed necessary prepossessing on the dataset to overcome data discrepancies, noise, and outliers to ensure the quality of discovered patterns as described in \cite{imran2023enhancing}.

From this dataset, we only use reviews in English with at least one sentiment word as our goal is to test the compositionality or non-compositionality of SA (§ Introduction). We thus discard 350 reviews in English that do not contain any polarity expression at all, so our evaluation is conducted on the remaining reviews that do contain subjectivity. If the review is a complex sentence and contains more than one polarity expression, e.g. ``This hotel is expensive, but the staff is nice'', it is assigned a list of polarity values associated with each polarity expression such as [negative, positive] \cite{barnes-etal-2022-semeval}. Then, the aggregate polarity for the review as a whole is computed as the majority value in that list, or a third polarity (neutral) if there is no majority value (e.g. [negative, positive]) \cite{imran2024syntax}. We thus have a task of predicting three polarity labels (neutral, positive, and negative) on this dataset \cite{imran2024syntax}. 

Negation poses an additional problem to the polarity prediction and our goal is thus to test compositional and non-compositional approaches dealing with negation. This is why we created another dataset just containing sentences with negation such as "Chocolate is not tasty". For this, we extracted a subset dataset containing reviews with negative words like "not" (see \cite{imran2024syntax} for the negation words in English). We call the dataset that contains all reviews expressing subjectivity "Data All" and the dataset with reviews containing negative words "Data Negation". In addition, we create a subset of the data that contains coordination to test the modification of our code in § Code Modification. We label this dataset as "Data Coordination." We use accuracy as our evaluation metric.

\section{Results and Discussion}

\subsection{Original vs. Modified Compositional Approach}

Table~\ref{tab:fr} shows that there is an improvement of three percent accuracy between the original code from \cite{kellert2023experimenting} and its modified version that captures coordination as presented in § Code Modification.

\begin{table}[h!]
\caption{Accuracy Results from Comparison between Original Code and Modified Code}
\begin{center}
\begin{tabular}{l l c}
\hline
\textbf{Method} & \textbf{Dataset} & \textbf{Accuracy} \\ [2pt]
\hline
    Original version from \cite{kellert2023experimenting} & Data Coordination & 0.71\\
    Modified version & Data Coordination & 0.74\\ [2pt]
\hline
\end{tabular}
\end{center}
\label{tab:fr}
\end{table}

\subsection{Compositional vs. Non-compositional Approaches}
    
Table~\ref{tab:fr2} shows a difference in accuracy of 9 percent between compositional and non-compositional approaches in the dataset containing all reviews (see compositional M. 0.80 vs. non-compos. M. 0.71). However, the distinction decreases with negative subjective statements (see compositional M. 0.72 vs. non-compositional M. 0.70).        

\begin{table}[h!]
\caption{Accuracy Results from Comparison between Compositional and Non-Compositional Approaches}
\begin{center}
\begin{tabular}{l l c}
\hline
\textbf{Method} & \textbf{Dataset} & \textbf{Accuracy} \\ [2pt]
\hline
Compositional dictionary-based & Data Negation & 0.72 \\
Compositional dictionary-based & Data All & 0.80 \\
Non-Compositional dictionary-based & Data Negation & 0.70 \\
Non-Compositional dictionary-based & Data All & 0.71 \\ [2pt]
\hline
\end{tabular}
\end{center}
\label{tab:fr2}
\end{table}

We also use a qualitative data analysis method, to compare the methods and their results. We subdivide the results into four conditions: false prediction by the compositional method, but correct prediction by the non-compositional method VADER (Condition 1). False prediction by both methods (Condition 2). False prediction by VADER, but correct prediction by the compositional method (Condition 3). Correct prediction by both methods (Condition 4). 

\begin{itemize}
\item \verb|Condition 1| : "However, this does not make up for the expense and lack of space."
\item \verb|Condition 2 | : "Room Tip : Best rooms are in another hotel, not there."
\item \verb|Condition 3| : "There was nothing that we did not like at this hotel."
\item \verb|Condition 4 | :"It is worn down, not clean and the whole hotel looks like a mess." 
\end{itemize}

Compositional approaches have issues with the scope of negation of certain semantic word classes that are considered propositional verbs like "make up" or "explain" that usually take a proposition or a sentence as their argument (see Cond. 1). In the example representing Cond.1, the arguments of the verb 'make up' are conjoined noun phrases "the expense" and "lack of space". However, their interpretation is a concealed proposition: "this does not make up for the expense and lack of space" means "this does not explain why the hotel is expensive and why there is a lack of space." Consequently, the negation does not scope over the sentence including nominal arguments, but over the main verb "make up". This case represents a mismatch between the syntactic structure that predicts the scope of negation over the clause including nominal arguments and semantic interpretation of the sentence where negation has scope over the main verb only. This is why our compositional approach does not correctly capture the polarity prediction of the sentence.
 Another problem is anaphoric and deictic expressions like "other", "here", etc. We see in the example representing Condition 2 that both methods have difficulties in capturing anaphoric and deictic references to targets of subjective statements (see Cond.2). In more recent approaches of SA, targets play an important role in polarity prediction \cite {barnes-etal-2022-semeval}. Targets as well as other concepts need to be adapted to compositional approaches in the future.   
The non-compositional method has problems in predicting polarity if the negation does not show a close proximity to the sentiment word as in the example representing Condition 3. The dependency parser, however, correctly interprets the negative argument "nothing" as the object argument of the verb "like", thereby correctly predicting the scope of negation. The double negation in the example leads to a positive interpretation and hence to the positive polarity. The non-compositional approach seems to have trouble with non-proximal negative words and the correct interpretation of double negation in contrast to the compositional approach \cite{Vilares2017, kellert2023experimenting}.  
Both methods (compositional and non-compositional) correctly predict the polarity of a sentence, whenever the scope of negation coincides with the linear order of the negation word as in "not clean" (see Cond.4). In the given example of Cond.4, the negation is close to the sentiment word "clean", hence the non-compositional approach can correctly predict the polarity of "not clean". Even though the compositional approach is independent of the linear order of negation, it correctly predicts the scope of negation in "not clean" as the negation word "not" is a modifier of the adjective "clean". Consequently, both approaches predict correctly the polarity orientation of the given example.

We have argued that Focus Analysis (FA) and Sentiment Analysis (SA) are strongly related, and the results from compositional approaches in SA based on sentiment datasets can be used as evaluation metrics for compositional approaches in FA. FA provides context and clarity about what or who is being discussed, while SA captures the emotional tone or attitude towards that focus. Their integration leads to a more accurate and nuanced understanding of texts, especially in complex cases involving multiple entities or topics. We have adjusted the head-child dependency relation using the English UD parser from \cite{kellert2023experimenting} to control the scope of negation, coordination, and modification. Furthermore, we suggested using specific datasets that are better suited for evaluating compositional approaches in SA than previous evaluations that do not consider particular datasets, such as short reviews and data with negation.

We evaluated both approaches (our compositional approach and the non-compositional approach from VADER) by the accuracy of polarity prediction performed on two datasets (Data All and Data Negation). Our results show that the compositional method we adapted has much higher accuracy than the non-compositional approach of VADER on the dataset "Data All" and almost similar accuracy on the second dataset "Data Negation" (with only a two-percent difference). The lack of a big distinction between compositional and non-compositional approaches in the dataset "Data Negation" can be attributed to a relatively high match between the linear order of negation and the sentiment word, such as in "not clean," and the scope of negation in the dataset. This is confirmed by examples showing mismatches between the linear order of negation and the sentiment word and the scope of negation, as discussed in the qualitative data analysis. These analyses show better performance of compositional approaches whenever the negation is correctly parsed as a modifier by the UD parser, despite being distant from the sentiment word (see Cond.3). As such mismatches are relatively rare in our dataset, the improvement of the compositional approach does not stand out.

\section{Conclusion}
To conclude, compositional approaches show certain advantages compared to non-compositional approaches, but there are still issues to address in future research, such as the scope of negation, anaphoric and deictic expressions, and the integration of targets into polarity prediction. In our previous work, we demonstrated the extent to which polarity prediction of compositional approaches depends on the selection of dictionaries \cite{kellertdict}. Most sentiment dictionaries ignore the issue of word ambiguity. For instance, the word old has a negative score in SO-CAL, but it can be used in contexts where it does not convey negative sentiment, such as "old tradition," "old friend," or "old town." Several approaches have addressed Word Sense Disambiguation (WSD), including WSD for lexicon-based approaches for SA \cite{VilAloGomDocEng2013, VASHISHTHA2019112834}. Even though the scope of our paper is not WSD but the evaluation of compositionality, WSD affects the evaluation of accuracy in polarity prediction of compositional approaches and, therefore, cannot be completely ignored. We will include the issue of WSD in future evaluations of compositional and non-compositional approaches. Additionally, a large dataset covering various focus expressions is necessary to ensure that a compositional approach can address all types of focus expressions, not just sentiment expressions.

\section{Limitations and Future Work}
We don't use any optimization or other task-specific adjustments in our experiments to try to increase the polarity prediction task's accuracy. This is because our primary goal in this study is not to increase the precision of the sentiment analysis, but instead to enhance Syntactic Parsing's temporal constraints so that it can benefit from its explainability and transparency compared to strictly supervised methods. Our method's use of a single English dataset to gauge accuracy in the polarity prediction task is another drawback. This is due to the fact that our method needs language-specific sentiment dictionaries in order to recognize polarity-shifting components like intensification and negation. In future, we will obtain these resources to test and evaluate our study in other languages too.

\section{Acknowledgments}

We acknowledge the European Research Council (ERC), which has funded this research under the Horizon Europe research and innovation programme (SALSA, grant agreement No 101100615), ERDF/MICINN-AEI (SCANNER-UDC, PID2020-113230RB-C21), Xunta de Galicia (ED431C 2020/11), and Centro de Investigación de Galicia ‘‘CITIC’’, funded by Xunta de Galicia and the European Union (ERDF - Galicia 2014–2020 Program), by grant ED431G 2019/01, LATCHING (PID2023-147129OB-C21) funded by MICIU/AEI/10.13039/501100011033 and ERDF, EU" and also "TSI-100925-2023-1 funded by Ministry for Digital Transformation and Civil Service and “NextGenerationEU” PRTR.

% Add the bibliography
\renewcommand{\bibname}{References} % For report or book class
\bibliographystyle{plainnat} % Specify the style (plainnat, abbrvnat, etc.)
\bibliography{references}   % Name of your .bib file (without .bib extension)

\begin{thebibliography}{29}
\providecommand{\natexlab}[1]{#1}
\providecommand{\url}[1]{\texttt{#1}}
\expandafter\ifx\csname urlstyle\endcsname\relax
  \providecommand{\doi}[1]{doi: #1}\else
  \providecommand{\doi}{doi: \begingroup \urlstyle{rm}\Url}\fi

\bibitem[Agerri et~al.(2013)Agerri, Cuadros, Gaines, and Rigau]{Agerri2013}
Rodrigo Agerri, Montse Cuadros, Sean Gaines, and German Rigau.
\newblock {OpeNER: Open polarity enhanced named entity recognition.}
\newblock In \emph{Sociedad Espa{\~{n}}ola para el Procesamiento del Lenguaje Natural}, volume~51, pages 215--218, 2013.

\bibitem[{\'A}lvarez-Carmona et~al.(2023){\'A}lvarez-Carmona, D{\'i}az-Pacheco, Aranda, Rodr{\'\i}guez-Gonz{\'a}lez, Bustio-Mart{\'i}nez, Mu{\~n}is-S{\'a}nchez, Pastor-L{\'o}pez, and S{\'a}nchez-Vega]{alvarez2023overview}
Miguel~{\'A} {\'A}lvarez-Carmona, {\'A}ngel D{\'i}az-Pacheco, Ram{\'o}n Aranda, Ansel~Y Rodr{\'\i}guez-Gonz{\'a}lez, L{\'a}zaro Bustio-Mart{\'i}nez, Victor Mu{\~n}is-S{\'a}nchez, A~Pastor Pastor-L{\'o}pez, and Fernando S{\'a}nchez-Vega.
\newblock Overview of rest-mex at iberlef 2023: Research on sentiment analysis task for mexican tourist texts.
\newblock \emph{Procesamiento del Lenguaje Natural}, 71, 2023.

\bibitem[Barnes et~al.(2022)Barnes, Oberlaender, Troiano, Kutuzov, Buchmann, Agerri, {\O}vrelid, and Velldal]{barnes-etal-2022-semeval}
Jeremy Barnes, Laura Oberlaender, Enrica Troiano, Andrey Kutuzov, Jan Buchmann, Rodrigo Agerri, Lilja {\O}vrelid, and Erik Velldal.
\newblock {S}em{E}val 2022 task 10: Structured sentiment analysis.
\newblock In \emph{Proceedings of the 16th International Workshop on Semantic Evaluation (SemEval-2022)}, pages 1280--1295, Seattle, United States, July 2022. Association for Computational Linguistics.
\newblock \doi{10.18653/v1/2022.semeval-1.180}.
\newblock URL \url{https://aclanthology.org/2022.semeval-1.180}.

\bibitem[Beck(2006)]{beck2006}
Sigrid Beck.
\newblock Intervention effects follow from focus interpretation.
\newblock \emph{Natural Language Semantics}, 14\penalty0 (1):\penalty0 1--56, 2006.

\bibitem[Brooke et~al.(2009)Brooke, Tofiloski, and Taboada]{BroTofTab2009a}
J.~Brooke, M.~Tofiloski, and M.~Taboada.
\newblock {Cross-Linguistic Sentiment Analysis: From English to Spanish}.
\newblock In \emph{Proceedings of RANLP 2009, Recent Advances in Natural Language Processing}, pages 50--54, Bovorets, Bulgaria, September 2009.

\bibitem[Chafe(1976)]{chafe1976}
Wallace~L. Chafe.
\newblock Givenness, contrastiveness, definiteness, subjects, topics, and point of view.
\newblock In Charles~N. Li, editor, \emph{Subject and Topic}, pages 25--55. Academic Press, New York, 1976.

\bibitem[Chierchia(2004)]{chierchia2004}
Gennaro Chierchia.
\newblock Scalar implicatures, polarity phenomena and the syntax/pragmatics interface.
\newblock In Adriana Belletti, editor, \emph{Structures and Beyond}, page [insert page range]. Oxford University Press, Oxford, 2004.

\bibitem[G{\'o}mez-Rodr{\'\i}guez et~al.(2024)G{\'o}mez-Rodr{\'\i}guez, Imran, Vilares, Solera, and Kellert]{gomez2024dancing}
Carlos G{\'o}mez-Rodr{\'\i}guez, Muhammad Imran, David Vilares, Elena Solera, and Olga Kellert.
\newblock Dancing in the syntax forest: fast, accurate and explainable sentiment analysis with salsa.
\newblock \emph{arXiv preprint arXiv:2406.16071}, 2024.

\bibitem[Graf and Marcinek(2014)]{DBLP:conf/acl-cmcl/GrafM14}
Thomas Graf and Bradley Marcinek.
\newblock Evaluating evaluation metrics for minimalist parsing.
\newblock In Vera Demberg and Timothy O'Donnell, editors, \emph{Proceedings of the Fifth Workshop on Cognitive Modeling and Computational Linguistics, CMCL@ACL 2014, Baltimore, Maryland, USA, June 26, 2014}, pages 28--36. Association for Computational Linguistics, 2014.
\newblock \doi{10.3115/v1/W14-2004}.
\newblock URL \url{https://doi.org/10.3115/v1/W14-2004}.

\bibitem[Hutto and Gilbert(2014)]{Hutto_Gilbert_2014}
C.~Hutto and Eric Gilbert.
\newblock Vader: A parsimonious rule-based model for sentiment analysis of social media text.
\newblock \emph{Proceedings of the International AAAI Conference on Web and Social Media}, 8\penalty0 (1), May 2014.
\newblock \doi{10.1609/icwsm.v8i1.14550}.
\newblock URL \url{https://ojs.aaai.org/index.php/ICWSM/article/view/14550}.

\bibitem[Imran and Ahmad(2023)]{imran2023enhancing}
Muhammad Imran and Adnan Ahmad.
\newblock Enhancing data quality to mine credible patterns.
\newblock \emph{Journal of Information Science}, 49\penalty0 (2):\penalty0 544--564, 2023.

\bibitem[Imran et~al.(2024)Imran, Kellert, and G{\'o}mez-Rodr{\'\i}guez]{imran2024syntax}
Muhammad Imran, Olga Kellert, and Carlos G{\'o}mez-Rodr{\'\i}guez.
\newblock A syntax-injected approach for faster and more accurate sentiment analysis.
\newblock \emph{arXiv preprint arXiv:2406.15163}, 2024.

\bibitem[Jacobs(1984)]{jacobs1984}
Joachim Jacobs.
\newblock \emph{Fokus und Skalen}.
\newblock Niemeyer, Tübingen, 1984.

\bibitem[Kellert et~al.(2023)Kellert, Zaman, Matlis, and G{\'o}mez-Rodr{\'\i}guez]{kellert2023experimenting}
Olga Kellert, Mahmud~Uz Zaman, Nicholas~Hill Matlis, and Carlos G{\'o}mez-Rodr{\'\i}guez.
\newblock Experimenting with ud adaptation of an unsupervised rule-based approach for sentiment analysis of mexican tourist texts.
\newblock \emph{arXiv preprint arXiv:2309.05312}, 2023.

\bibitem[Kellert et~al.(2024)Kellert, Gómez-Rodríguez, and Uz~Zaman]{kellertdict}
Olga Kellert, Carlos Gómez-Rodríguez, and Mahmud Uz~Zaman.
\newblock Unveiling factors influencing judgment variation in sentiment analysis with natural language processing and statistics.
\newblock \emph{PLOS ONE}, 19\penalty0 (5):\penalty0 1--19, 05 2024.
\newblock \doi{10.1371/journal.pone.0304201}.
\newblock URL \url{https://doi.org/10.1371/journal.pone.0304201}.

\bibitem[Kellert(2015)]{kellert2015}
Oliver Kellert.
\newblock \emph{Interrogative und Exklamative: Syntax und Semantik von multiplen wh-Elementen im Französischen und Italienischen}.
\newblock PhD thesis, University of Göttingen, Berlin, 2015.
\newblock Linguistische Arbeiten 560.

\bibitem[Krifka(1991)]{krifka1991}
Manfred Krifka.
\newblock A compositional semantics for multiple focus constructions.
\newblock In Steven Moore and Adam~Zachary Wyner, editors, \emph{Proceedings of the First Semantics and Linguistic Theory Conference (SALT 1)}, pages 127--158. Cornell University, Ithaca, NY, 1991.

\bibitem[Lambrecht(1994)]{lambrecht1994}
Knud Lambrecht.
\newblock \emph{Information Structure and Sentence Form: A Theory of Topic, Focus, and the Mental Representations of Discourse Referents}.
\newblock Cambridge University Press, Cambridge, 1994.

\bibitem[Müller et~al.(2021)Müller, Abeillé, Borsley, and Koenig]{Müller2021}
Stefan Müller, Anne Abeillé, Robert~D. Borsley, and Jean-Pierre Koenig, editors.
\newblock \emph{{Head} {Driven} {Phrase} {Structure} {Grammar}}.
\newblock Number~9 in Empirically Oriented Theoretical Morphology and Syntax. Language Science Press, Berlin, 2021.
\newblock \doi{10.5281/zenodo.5543318}.

\bibitem[Qi et~al.(2018)Qi, Dozat, Zhang, and Manning]{qi-etal-2018-universal}
Peng Qi, Timothy Dozat, Yuhao Zhang, and Christopher~D. Manning.
\newblock {U}niversal {D}ependency parsing from scratch.
\newblock In \emph{Proceedings of the {C}o{NLL} 2018 Shared Task: Multilingual Parsing from Raw Text to Universal Dependencies}, pages 160--170, Brussels, Belgium, October 2018. Association for Computational Linguistics.
\newblock \doi{10.18653/v1/K18-2016}.
\newblock URL \url{https://aclanthology.org/K18-2016}.

\bibitem[Qi et~al.(2020)Qi, Zhang, Zhang, Bolton, and Manning]{qi-etal-2020-stanza}
Peng Qi, Yuhao Zhang, Yuhui Zhang, Jason Bolton, and Christopher~D. Manning.
\newblock {S}tanza: A python natural language processing toolkit for many human languages.
\newblock In \emph{Proceedings of the 58th Annual Meeting of the Association for Computational Linguistics: System Demonstrations}, pages 101--108, Online, July 2020. Association for Computational Linguistics.
\newblock \doi{10.18653/v1/2020.acl-demos.14}.
\newblock URL \url{https://aclanthology.org/2020.acl-demos.14}.

\bibitem[Roberts(1996)]{roberts1996}
Craige Roberts.
\newblock Information structure in discourse: Towards an integrated formal theory of pragmatics.
\newblock In Jeong-Hye Yoon and Andreas Kathol, editors, \emph{OSU Working Papers in Linguistics}, volume~49, pages 91--136. The Department of Linguistics of Ohio State University, 1996.

\bibitem[Rooth(1985)]{rooth1985}
Mats Rooth.
\newblock \emph{Association with Focus}.
\newblock PhD thesis, University of Massachusetts Amherst, 1985.

\bibitem[Taboada et~al.(2011)Taboada, Brooke, Tofiloski, Voll, and Stede]{Lexicon-BasedMethods}
M.~Taboada, J.~Brooke, M.~Tofiloski, K.~Voll, and M.~Stede.
\newblock {Lexicon-based methods for sentiment analysis}.
\newblock \emph{Computational Linguistics}, 37\penalty0 (2):\penalty0 267--307, 2011.
\newblock ISSN 0891-2017.

\bibitem[Vashishtha and Susan(2019)]{VASHISHTHA2019112834}
Srishti Vashishtha and Seba Susan.
\newblock Fuzzy rule based unsupervised sentiment analysis from social media posts.
\newblock \emph{Expert Systems with Applications}, 138:\penalty0 112834, 2019.
\newblock ISSN 0957-4174.
\newblock \doi{https://doi.org/10.1016/j.eswa.2019.112834}.
\newblock URL \url{https://www.sciencedirect.com/science/article/pii/S0957417419305366}.

\bibitem[Vilares et~al.(2013)Vilares, Alonso, and G\'{o}mez-Rodr\'{\i}guez]{VilAloGomDocEng2013}
David Vilares, Miguel~A. Alonso, and Carlos G\'{o}mez-Rodr\'{\i}guez.
\newblock {Supervised polarity classification of Spanish tweets based on linguistic knowledge}.
\newblock In \emph{DocEng'13. Proceedings of the 13th ACM Symposium on Document Engineering}, pages 169--172, Florence, Italy, September 2013. ACM.
\newblock ISBN 978-1-4503-1789-4.

\bibitem[Vilares et~al.(2017)Vilares, G{\'{o}}mez-Rodr{\'{\i}}guez, and Alonso]{Vilares2017}
David Vilares, Carlos G{\'{o}}mez-Rodr{\'{\i}}guez, and Miguel~A. Alonso.
\newblock Universal, unsupervised (rule-based), uncovered sentiment analysis.
\newblock \emph{Knowledge-Based Systems}, 118:\penalty0 45--55, feb 2017.
\newblock \doi{10.1016/j.knosys.2016.11.014}.
\newblock URL \url{https://doi.org/10.10162Fj.knosys.2016.11.014}.

\bibitem[von Stechow(1991)]{vonstechow1991}
Arnim von Stechow.
\newblock Focusing and backgrounding operators.
\newblock In Werner Abraham, editor, \emph{Discourse Particles}, pages 37--84. John Benjamins, Amsterdam, 1991.

\bibitem[Zeman et~al.(2023)Zeman, Nivre, Abrams, Ackermann, Aepli, Aghaei, et~al.]{11234/1-5150}
Daniel Zeman, Joakim Nivre, Mitchell Abrams, Elia Ackermann, No{\"e}mi Aepli, Hamid Aghaei, et~al.
\newblock Universal dependencies 2.12, 2023.
\newblock URL \url{http://hdl.handle.net/11234/1-5150}.
\newblock {LINDAT}/{CLARIAH}-{CZ} digital library at the Institute of Formal and Applied Linguistics ({{\'U}FAL}), Faculty of Mathematics and Physics, Charles University.

\end{thebibliography}
\section{Appendices}
%\appendix

\subsection{Formal details of FA}
\label{subsec:formal_details_FA}

A key component of Rooth's focus theory is the focus operator $\sim$, which takes propositional scope over the entire sentence and integrates the focus interpretation with the context. The scope of the operator is represented in [$\sim$C] as follows: $\sim$ C[{\itshape John} left]. The operator $\sim$ "resets" the focus value to an ordinary value, ensuring that only the relevant alternatives are considered, thereby preventing an overload of uninterpreted alternatives (for modifications and extensions of this theory, see Beck 2006). Rooth's assumption that every sentence with a focus expression implies a (covert) focus operator at a certain level of representation, usually at the sentential level or scope, is shared by other linguists (see Jacobs 1984, Chierchia 2004).

Rooth assumes a compositional analysis of focus. Compositionality refers to how the meaning of a sentence is built up from the meanings of its parts, specifically addressing how focus affects this process. The compositionality is captured by the function application rule. According to Beck's (2006) extension of Rooth's focus analysis, the function application rule applies to both the ordinary and focus interpretations, which are derived in a parallel manner. The function application rule states that if X is a phrase composed of two parts Y and Z, then the interpretation of X under an assignment function g (which assigns values to variables) is the result of applying the interpretation of Y to the interpretation of Z under g. 

Formally, this is represented as:
\[
[[X]]_g = [[Y]]_g([[Z]]_g)
\]

Additionally, when considering focus, another function h is introduced to handle the focus interpretation. Thus, the interpretation of X under both g and h is:
\[
[[X]]_{g,h}=[[Y]]_{g,h}([[Z]]_{g,h})
\]

To illustrate this with an example, consider the sentence "{\itshape John} left". The ordinary interpretation and focus interpretation are derived compositionally in parallel.

For the ordinary interpretation:

\begin{align*}
\text{left}_{g,h}(\textit{John}, h) &= \lambda x. \text{left}(x)(\textit{John}, h) = \text{left}(h(1)) \\
\text{left}_{g,h}(\textit{John}, h) &= \lambda x. \text{left}(x)(\text{John}_{\text{Focus},g,h}) = \text{left}(h(1))
\end{align*}

Here, \textit{John}$_g$ is interpreted as the value assigned to \textit{John} by the function $g$, and \textit{left}$_g$ is the function that applies to this value, resulting in \textit{left}$(g(1))$.

For the focus interpretation:

%\begin{document}

\begin{align*}
\text{left}_{g,h}(\textit{John}, h) &= \lambda x. \text{left}(x)(\textit{John}, h) \\
                                    &= \text{left}(h(1)) \\
\text{left}_{g,h}(\textit{John}, h) &= \lambda x. \text{left}(x)(\text{John}_{\text{Focus},g,h}) \\
                                    &= \text{left}(h(1))
\end{align*}

In this case, \textit{John}$_{g,h}$ is interpreted according to the focus function $h$, and \textit{left}$_g$ applies to this focused value, resulting in \textit{left}$(h(1))$.

By maintaining separate functions g and h for ordinary and focus interpretations respectively, Beck (2006) ensures that both interpretations are systematically and compositionally derived, reflecting how focus elements influence the meaning of sentences within a context. This dual approach allows for a precise and structured handling of how focus affects interpretation, integrating Rooth's \cite{rooth1985} insights into a formal semantic framework.  

\subsection{Formal details of SA}
\label{subsec:formal_details_SA}

\begin{figure*}
  \centering
  \includegraphics[width=\linewidth]{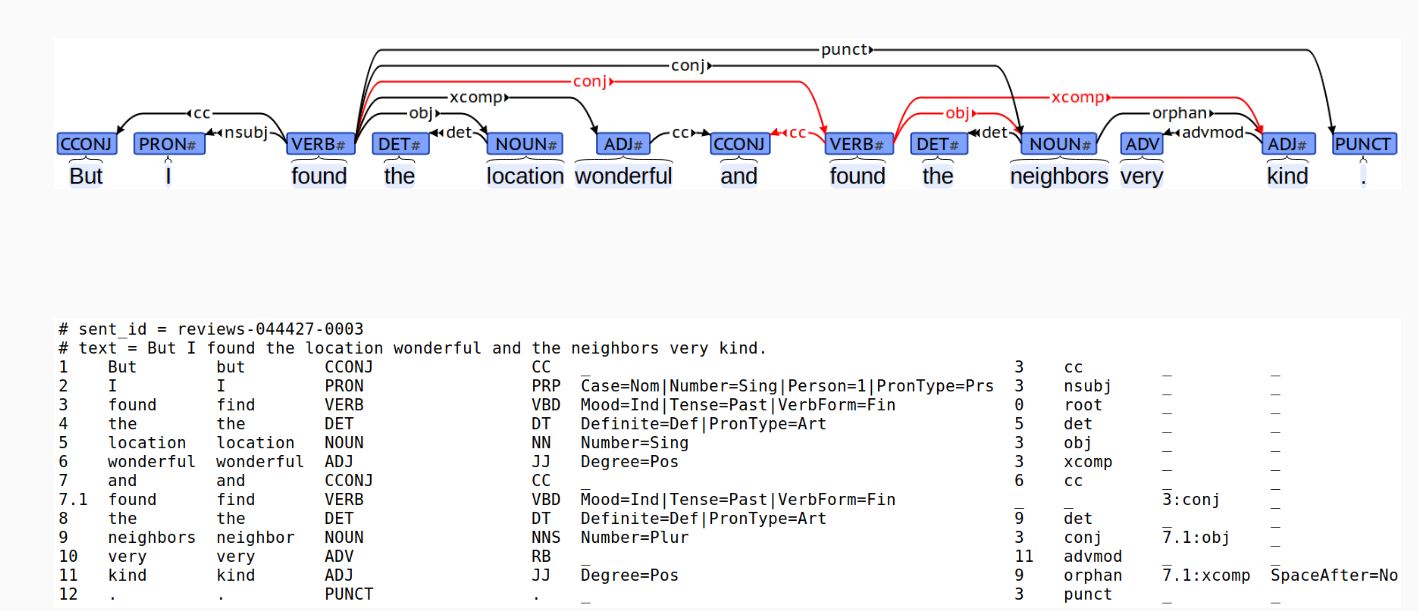}
  \caption{UD formalism.(\protect\url{http://universaldependencies.org/eacl17tutorial/infrastructure.pdf}).}
  \label{fig:UD}
\end{figure*}

Following are formal details of some compositional approaches in SA. The authors in \cite {kellert2023experimenting} suggest an adaptation of an unsupervised compositional and recursive approach in SA (Vilares et al., 2017 \cite{Vilares2017} ) to the Universal Dependencies (UD) formalism \cite {11234/1-5150}, as it has since become the de facto standard for multilingual dependency parsing. Figure ~\ref{fig:UD} shows a dependency structure for an English sentence and a CoNLL-U Format which represents word lines containing the annotation of a word/token with respect to various linguistic properties such as part of speech (POS), lemma, dependency relation of the word to its head, morphological features, etc. The dependency structure and linguistic properties of word/tokens as in CoNLL-U Format are an integral part of UD.

The authors in \cite {kellert2023experimenting}) used Stanza, which is a natural language toolkit based on UD-formalism that provides a basic analysis of the input text such as lemmatization, part-of-speech (POS) and dependency parsing (Peng Qi et al., 2020 \cite {qi-etal-2020-stanza}). The dependency parser is based on UD parser from Qi et al. 2018 \cite {qi-etal-2018-universal}. To demonstrate the approach in \cite {kellert2023experimenting}, let's consider a Spanish example {\itshape No es excelente} `It is not excellent' and the associated dictionary entries with token ids, text, lemma, POS (`upos'), morphological features (`feats'), head ids and dependency relations (`deprel'): 

\begin{itemize}

\item \verb|first word| : {`id': 1, `text': `no', `lemma': `no', `upos': `ADV', `feats': `Polarity=Neg', `head': 3, `deprel': `advmod'} 
\item \verb|second word| : {`id': 2, `text': `es', `lemma': `ser', `upos': `AUX', `feats': `feats': `Mood=Ind|...', `head': 3, `deprel': `cop'} 
\item \verb|third word| : {`id': 3, `text': `excelente', `lemma': `excelente', `upos': `ADJ', `feats': `Number=Sing', `head': 0, `deprel': `root'}

\end{itemize}

Head ids and dependency relations play an important role as they provide information about the syntactic relation of words and the hierarchical structure of the sentence. Head ids contain information about parent-child relations. Take for instance, the negation word {\itshape no} and the copular word {\itshape es} in the previous example, which have {\itshape excelente} as their head. This means that the word {\itshape excelente} is the highest node and the children {\itshape no} and {\itshape es} are the lowest nodes in the structure. This head-child relation can be used to define the scope of negation. If the negation is a child of a sentiment word as its head, the polarity of the sentiment word needs to be shifted.   

In order to be able to calculate the polarity score of a sentence, they perform several steps that can be described in a nutshell as follows:

\begin{itemize}
\item \verb|Step 1| : Find sentiment words in the input text and assign polarity scores to the sentiment words
\item \verb|Step 2| : Create a dictionary of head ids and their correspondent children ids 
\item \verb|Step 3| : Identify target words that influence the sentiment word such as negation 
\item \verb|Step 4| : Calculate the polarity score for the input sentence 

\end{itemize}

Let us illustrate these steps by looking at the given Spanish example. First, the authors identify the sentiment word {\itshape excelente} in the input text and add new entries to the dictionary associated with this word, namely the { elementType: `count'} and the polarity score or {`elementScore': 5}. They use the dictionaries by SO-CAL for Spanish \cite {Lexicon-BasedMethods}, \cite {Vilares2017}, in which the polarity score for sentiment words ranges from -5 (the most negative) to +5 (the most positive).

\begin{itemize}
\item \verb|Sentence| : {\itshape No es excelente} `It's not excellent'
\item \verb|Step 1| : label sentiment words
\item \verb|dictionary of the sentiment word| : {`id': 3, `text': `excelente', `lemma': `excelente', `upos': `ADJ', `feats': `Number=Sing', `head': 0, `deprel': `root', `elementType': `count', `elementScore': 5}

\end{itemize}

\begin{figure}
  \centering
  \includegraphics[width=\linewidth]{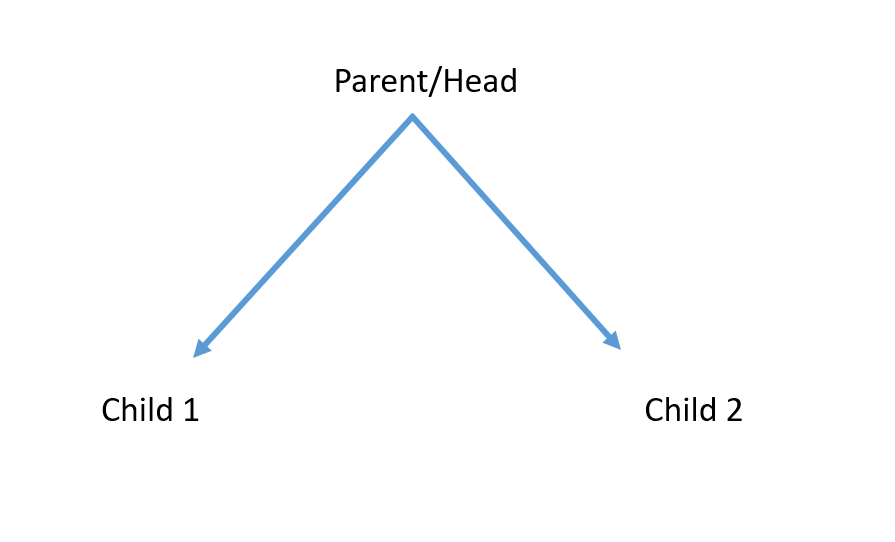}
  \caption{Example of a head-child tree branch}
  \label{fig:head-child}
\end{figure}

Step 2 consists of creating a dictionary with head ids as keys and a list of children as a key value in order to find potential polarity shifters or target words such as negation and modification. Each key-value pair of this dictionary represents a head-child tree branch as represented in Figure ~\ref{fig:head-child}.

In the UD-formalism, the head id 0 and its child represent the highest tree branch and the child of the head id 0 and its children represent the second highest branch. In the given example, the second-highest branch is also the lowest branch: 

\begin{itemize}
\item \verb|Sentence| : {\itshape No es excelente}
\item \verb|Step 2| : Create a dictionary with heads as keys and their correspondent children as values
\item \verb|head-child-dictionary| : {3: [1, 2], 0: [3]}

\end{itemize}

Head-child branches represent an important unit in compositional approaches to FA and SA. In FA, the head represents the function in Functional Application (FA) applied to arguments of the sentence (see § Compositional approaches to Focus and Sentiment Analysis). In SA, head-child branches have been used to identify target words that can shift, weaken or strengthen the polarity of sentiment words \cite {kellert2023experimenting}. 

Step 3 consists of identifying target words that can modify the sentiment word identified in step 1. To achieve this goal, the authors loop through branches upwards and check for a sentiment word, negation and/or modification in the same branch. For this, they calculate the order of branches from the lowest to the highest branch associated with a sentence. In the given sentence example {\itshape no es excelente}, the sentiment word and negation are in the same branch. 

Step 4 consists of calculating the polarity score for each branch upwards by applying the formula for the calculation of the polarity score in (1) from Vilares et al. 2017 \cite {Vilares2017}, where the variable a equals the elementScore of a sentiment word such as {\itshape excelente}, the variable b equals a value that depends on the strength of the intensifier such as {\itshape muy} taken from a list of intensifiers and negation has a score of -4 or +4 depending on the positive or negative value of a:

\begin{itemize}
\item \verb|Step 4| : Calculate the polarity score for the branch {\itshape {3: [1, 2]}}
\begin{equation}
  \ a *(1 + b)+(sign(a)*-4) = polarity score
\end{equation}
\end{itemize}

According to the formula in (1), the polarity score for the lowest branch 3:[1, 2] equals 1, if we calculate 5*(1+0)-4. As the highest branch simply expresses an identity relation between the root and the head of the previous branch, the polarity score remains the same, namely 1, and the calculation finishes with the highest branch. The authors take the polarity score of the highest branch ("top branch") to be the final result for the polarity calculation. 

\begin{figure}
  \centering
  \includegraphics[width=\linewidth]{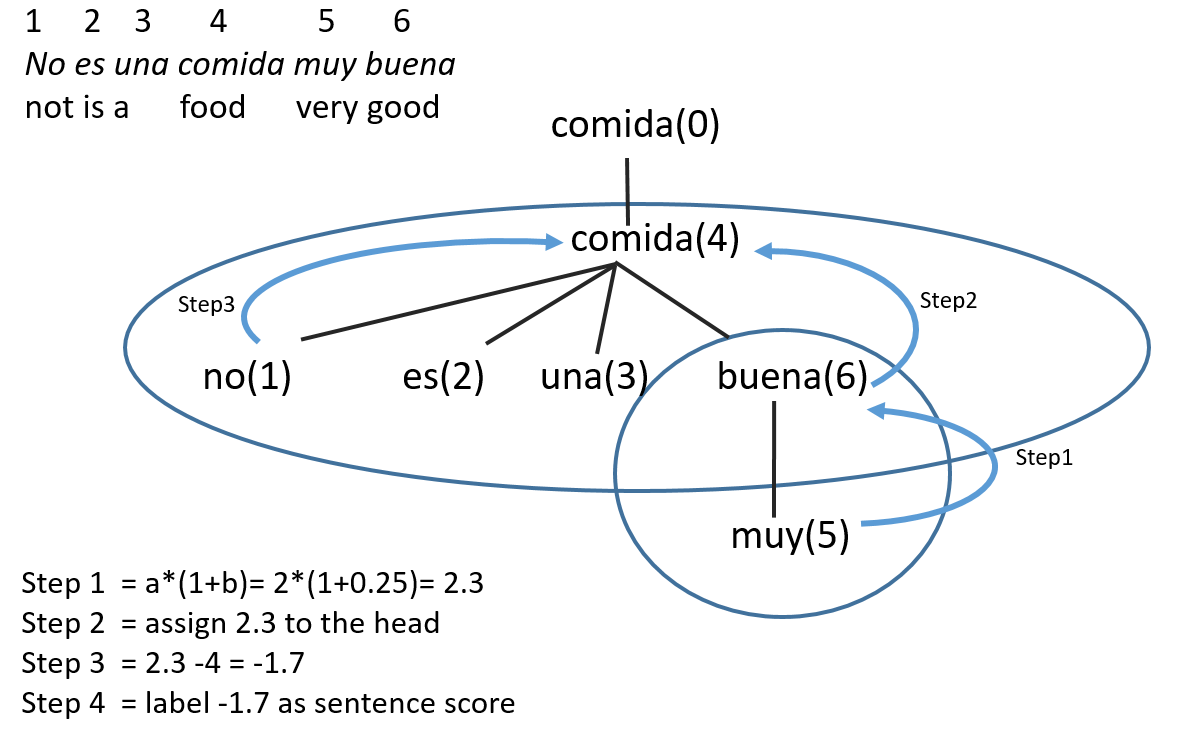}
  \caption{Example with several branches}
  \label{fig:nodes}
\end{figure}

They also discuss an example with more branches in Figure ~\ref{fig:nodes}. 

First, the intensification of the adjective {\itshape buena} `good' is computed by the intensifier {\itshape muy} `very'. The score of the intensifier {\itshape muy} is 0.25 \cite {Vilares2017}. The result for this calculation is 2*(1+0.25)=2.3. This score is assigned to the nominal head {\itshape comida} `food' as the result of the nominal modification. Collecting information from the lowest branch and bringing it up to the highest branch (e.g. nominal phrase) is a common step in formal grammars such as Head-driven Phrase Structure Grammar (HPSG) \cite {Müller2021} or Minimalist Grammar \cite {DBLP:conf/acl-cmcl/GrafM14}. As the negation is a child of the nominal head with a polarity score 2.3, the negation has scope over the nominal head. As a result, the polarity score 2.3 is substracted -4 and the output of the calculation is -1.7.

The calculation finishes with the highest branch, which expresses an identity relation between the root and its child. The calculation steps are summarized as follows: 

\begin{itemize}
\item \verb|Sentence| : {\itshape No es una comida muy buena} `It's not a very good food'
\item \verb|polarity score of the lowest branch | : 2 * (1 + 0.25) = 2.3 
\item \verb|polarity score of the higher branch | : 2.3 -4 = -1.7
\item \verb|polarity score of the highest branch | : -1.7 (final polarity score) 
\end{itemize}

\section{Online Resources}

The files used for the experiment and the updated code for the compositional SA are available on GitHub.

\begin{itemize}
\item \href{https://github.com/olga-kel/compositional-sentiment}{GitHub}

\end{itemize}

\end{document}